\documentclass{article}

\usepackage[preprint, nonatbib]{neurips_2024}

\usepackage[utf8]{inputenc} 
\usepackage[T1]{fontenc}    
\usepackage{hyperref}       
\usepackage{url}            
\usepackage{booktabs}       
\usepackage{amsfonts}       
\usepackage{nicefrac}       
\usepackage{microtype}      
\usepackage{xcolor}         

\usepackage{graphicx}
\usepackage{amsmath}
\usepackage{caption}
\usepackage{wrapfig}
\usepackage{hyperref}
\usepackage{subfigure}

\title{LLM Surgery: Efficient Knowledge Unlearning and Editing in Large Language Models}

\author{%
  Akshaj Kumar Veldanda\thanks{Email: akv275@nyu.edu} \\
  New York University \\
  \And
  Shi-Xiong Zhang \\
  Capital One \\
  \And 
  Anirban Das \\
  Capital One \\
  \And 
  Supriyo Chakraborty \\
  Capital One \\
  \And Stephen Rawls \\
  Capital One \\
  \And Sambit Sahu \\
  Capital One \\
  \And Milind Naphade \\
  Capital One \\
}

\begin{document}

\maketitle

\begin{abstract}
Large language models (LLMs) have revolutionized various domains, yet their utility comes with significant challenges related to outdated or problematic knowledge embedded during pretraining. This paper addresses the challenge of modifying LLMs to unlearn problematic and outdated information while efficiently integrating new knowledge without retraining from scratch. Here, we propose LLM Surgery, a framework to efficiently modify LLM behaviour by optimizing a three component objective function that: (1) Performs reverse gradient on unlearning dataset (problematic and outdated information), (2) Performs gradient descent on the update dataset (new and updated information), and (3) Minimizes the KL divergence on the retain dataset (small subset of unchanged text), ensuring alignment between pretrained and modified model outputs. Due to the lack of publicly available datasets specifically tailored for our novel task, we compiled a new dataset and an evaluation benchmark\footnote{We intend to make the code and dataset publicly available at \href{https://anonymous.4open.science/r/llm\_surgery\_code-24E8/README.md}{link}.}. Using Llama2-7B, we demonstrate that LLM Surgery can achieve significant forgetting on the unlearn set, a 20\% increase in accuracy on the update set, and maintain performance on the retain set.

\textbf{Keywords:} unlearning, editing, LLM, dataset, and evaluation benchmark.  
\end{abstract}

\section{Introduction}
Large language models (LLMs) excel across various tasks~\cite{vilar2022prompting, zhu2023multilingual, zhang2023sentiment, wulf2024exploring}, but they rely on extensive internet-sourced data~\cite{xu2024survey, achiam2023gpt}, which often includes problematic content such as copyrighted material and personal information~\cite{zhao2022provably}. This can lead to LLMs inadvertently memorizing and generating such data~\cite{smith2023identifying}. Additionally, LLMs are limited by the temporal scope of their training data, lacking awareness of events or updates after training~\cite{lazaridou2021mind, agarwal2022temporal}, which risks generating outdated or incorrect information.
Retraining LLMs to address these issues is computationally expensive; for example, pretraining Llama-3 requires 7.7 million GPU hours and produces significant CO$_2$ emissions~\cite{llama3modelcard}.

To ensure LLMs remain relevant and beneficial, it is essential to develop methods that allow models to \textbf{(1)} unlearn problematic or outdated data, \textbf{(2)} incorporate new information, and \textbf{(3)} maintain performance on standard benchmarks. Without these capabilities, LLMs pose significant legal and ethical risks~\cite{rtbf}, as shown by lawsuits filed by content creators, such as the New York Times, against companies using copyrighted material for training~\cite{times2023sue}. As LLMs become more widespread, the risk of perpetuating outdated or legally sensitive content grows, leading to severe consequences~\cite{zhao2022provably}.

To address these challenges, we propose LLM Surgery, a framework for efficiently updating LLM knowledge to ensure it remains current, accurate, and legally compliant. Our key contributions are: \textbf{(1)} the development of an objective function that applies reverse gradient to unlearn data, gradient descent to new data, and minimizes KL divergence on retained data, ensuring alignment between pretrained and updated outputs (see~\autoref{sec: methodology}). This function operates in a continual-pretraining regime, eliminating the need for costly annotations and enabling scalable unlearning and updating; \textbf{(2)} the creation of a comprehensive dataset with unlearn, update, and retain sets for both training and evaluation (see~\autoref{sec: setup}); and \textbf{(3)} empirical results and activation pattern analysis showing significant forgetting on the unlearn set, a 20\% accuracy improvement on the update set, and stable performance on the retain set (see~\autoref{sec: results}).

\section{Proposed Methodology} 
\label{sec: methodology}
Given three datasets — the unlearn dataset ($\mathcal{D}^{\text{unl}}$), the update dataset ($\mathcal{D}^{\text{upd}}$), and the retain dataset ($\mathcal{D}^{\text{rtn}}$) — our goal is to modify a LLM, $\boldsymbol{\theta}^{\text{pre}}$, so that the resulting LLM, $\boldsymbol{\theta}^{\text{surgery}}$, meets the following criteria: (1) it unlearns specific information from $\boldsymbol{\theta}^{\text{pre}}$ based on $\mathcal{D}^{\text{unl}}$, (2) updates $\boldsymbol{\theta}^{\text{pre}}$ with new knowledge from $\mathcal{D}^{\text{upd}}$, and (3) maintains the same level of performance as $\boldsymbol{\theta}^{\text{pre}}$ on standard benchmark tasks using $\mathcal{D}^{\text{rtn}}$. To achieve these goals efficiently, we propose the LLM Surgery framework.

\paragraph{LLM Surgery Objective} optimizes a objective function, as defined in~\autoref{eqn:obj}. This function is designed to accomplish three key objectives: (1) \textbf{\textit{Unlearning}}: The first term in~\autoref{eqn:obj} encourages the model to unlearn specific information by performing gradient ascent on $\mathcal{D}^{\text{unl}}$. This ensures that the model parameters are optimized along the reverse gradient direction of the learned patterns in $\mathcal{D}^{\text{unl}}$, effectively erasing targeted knowledge, (2) \textbf{\textit{Updating}}: The second term facilitates learning of new information by applying gradient descent on the $\mathcal{D}^{\text{upd}}$. This ensures the model integrates the updated knowledge, and (3) \textbf{\textit{Retention}}: The third term minimizes the KL-divergence between the output distributions of $\boldsymbol{\theta}^{\text{surgery}}$ and $\boldsymbol{\theta}^{\text{pre}}$ on $\mathcal{D}^{\text{rtn}}$. This ensures that the model retains essential knowledge that should remain unaffected by LLM Surgery. Mathematically, it can be expressed as follows:

\begingroup
\small
\begin{equation}
    \label{eqn:obj}
    \begin{aligned}
    \mathcal{L} = & \underbrace{\frac{1}{|\mathcal{D}^{\text{unl}}|}\sum_{i=1}^{|\mathcal{D}^{\text{unl}}|} \frac{1}{|x^{\text{unl}}_{i}|} \sum_{t=2}^{|x^{\text{unl}}_{i}|} l\Big(P[ \cdot \mid x^{\text{unl}}_{i; <t}; \boldsymbol{\theta}^{\text{surgery}}], \mathbf{e}_{t}^{\text{unl}} \Big)}_{\tt{unlearn}}
    - \underbrace{\frac{1}{|\mathcal{D}^{\text{upd}}|}\sum_{i=1}^{|\mathcal{D}^{\text{upd}}|} \frac{1}{|x^{\text{upd}}_{i}|} \sum_{t=2}^{|x^{\text{upd}}_{i}|} l\Big(P[ \cdot \mid x^{\text{upd}}_{i; <t}; \boldsymbol{\theta}^{\text{surgery}}], \mathbf{e}^{\text{upd}}_{t}\Big)}_{\tt{update}} \\
    &- \underbrace{\frac{1}{|\mathcal{D}^{\text{rtn}}|}\sum_{i=1}^{|\mathcal{D}^{\text{rtn}}|} \frac{1}{|x^{\text{rtn}}_{i}|} \sum_{t=2}^{|x^{\text{rtn}}_{i}|} \text{KL}\Big(P[\cdot \mid x^{\text{rtn}}_{i; < t}; \boldsymbol{\theta}^{\text{surgery}}] \Big\| P[\cdot \mid x^{\text{rtn}}_{i; < t}; \boldsymbol{\theta}^{\text{pre}}]\Big)}_{\tt{retain}}   \hspace{9em} (1)
    \nonumber
    \end{aligned}
\end{equation}
\endgroup

In this equation, $i$ denotes the $i^{\text{th}}$ example in the dataset, where $1 \leq i \leq |\mathcal{D}|$, and $|\mathcal{D}|$ represents the size of the dataset. $x^{\text{unl}}_i \in \mathcal{D}^{\text{unl}}$, $x^{\text{upd}}_i \in \mathcal{D}^{\text{upd}}$, and $x^{\text{rtn}}_i \in \mathcal{D}^{\text{rtn}}$ represent sequences of tokens from the unlearning, update, and retain datasets, respectively. The notation $P[ \cdot \mid x_{i;<t}; \boldsymbol{\theta}]$ refers to the predicted probability distribution over the possible next tokens at position $t$, given the preceding tokens $x_{<t}$ and the model parameters $\boldsymbol{\theta}$. The vector $\mathbf{e}_t$ is the one-hot encoded representation of the true token at position $t$, and $l(\cdot, \cdot)$ denotes the loss function, typically cross-entropy.

\section{Experimental Setup} 
\label{sec: setup}
We conduct experiments using the Llama2-7B model~\cite{llama2} to evaluate the effectiveness of the LLM Surgery framework in unlearning information from $\mathcal{D}^{\text{unl}}$, updating the model with new knowledge from $\mathcal{D}^{\text{upd}}$, and retaining performance on $\mathcal{D}^{\text{rtn}}$. As a preliminary step, we continually-pretrain \textit{off-the-shelf} Llama2-7B model on both $\mathcal{D}^{\text{unl}}$ and $\mathcal{D}^{\text{rtn}}$. The purpose of this continual-pretraining phase is to ensure that the model possesses the knowledge of $\mathcal{D}^{\text{unl}}$ and $\mathcal{D}^{\text{rtn}}$ prior to performing LLM Surgery. We refer to this continually-pretrained model as $\boldsymbol{\theta}^{\text{pre}}$.

\paragraph{Baseline Model:}A naive approach would involve continually-pretraining \textit{off-the-shelf} Llama2-7B model using only the update dataset $\mathcal{D}^{\text{upd}}$ and the \textit{full} retain dataset $\mathcal{D}^{\text{rtn}}$. This model is referred to as $\boldsymbol{\theta}^{\text{baseline}}$. However, this approach is both computationally expensive and time-intensive, particularly because $\mathcal{D}^{\text{rtn}}$ is significantly larger than $\mathcal{D}^{\text{upd}}$. The challenge is to achieve the desired modifications efficiently without compromising the overall performance of the model.

\paragraph{LLM Surgery:}To overcome these challenges more efficiently, LLM Surgery leverages the general-purpose knowledge from $\mathcal{D}^{\text{rtn}}$ already embedded in $\boldsymbol{\theta}^{\text{pre}}$ to make targeted modifications, resulting in the modified model $\boldsymbol{\theta}^{\text{surgery}}$. Instead of starting from the \textit{off-the-shelf} Llama2-7B model (as done when training $\boldsymbol{\theta}^{\text{baseline}}$), LLM Surgery begins with $\boldsymbol{\theta}^{\text{pre}}$. This approach efficiently unlearns specific information using $\mathcal{D}^{\text{unl}}$, integrates new knowledge using $\mathcal{D}^{\text{upd}}$, and preserves performance on critical tasks using a \textit{small} subset of $\mathcal{D}^{\text{rtn}}$. Crucially, it avoids the high computational costs of full retraining from scratch\footnote{Pretraining a LLM from scratch is resource-intensive and time consuming. Hence, we begin with \textit{off-the-shelf} Llama2-7B model.} (or in our context, retraining the Llama2-7B model on both $\mathcal{D}^{\text{upd}}$ and the \textit{entire} $\mathcal{D}^{\text{rtn}}$ datasets as in the baseline model).

\subsection{Data}
Given the lack of publicly available datasets specifically tailored for our task, we compiled a novel dataset and evaluation benchmark. This benchmark includes multiple-choice question-answer (MCQA) examples generated using GPT-4\footnote{GPT-4 prompt to generate our evaluation benchmark is inspired from~\cite{tofu_benchmark} and we use a 40-60 split for the validation and test sets.}, which consist of questions, three answer choices, and labels indicating the correct choice. Accuracy is used as the evaluation metric. 
Below, we describe the construction of the dataset and characteristics of the evaluation benchmark:

\paragraph{Unlearn Dataset ($\mathcal{D}^{\text{unl}}$):} We derive the unlearning dataset from the original Wikipedia biographies in the WikiBio GPT-3 Hallucination Dataset~\cite{wikibio_dataset}, consisting of 238 \textit{true} biographies. This dataset is divided into two subsets: (1) a subset of 119 biographies (50k tokens) used to simulate problematic data, such as personal information or copyrighted material, and (2) a subset of 119 biographies (50k tokens) representing outdated information. To assess the effectiveness of the LLM Surgery framework to unlearn this content from $\boldsymbol{\theta}^{\text{pre}}$, we generate 2,400 MCQA examples based on the unlearn dataset and used them for evaluating $\boldsymbol{\theta}^{\text{surgery}}$'s performance on the unlearning data. 

\paragraph{Update Dataset ($\mathcal{D}^{\text{upd}}$):} The update dataset is constructed by generating 119 fictitious biographies (80k tokens) in Wikipedia style using GPT-4~\cite{gpt4}. These biographies correspond to the same 119 subjects in the outdated subset of the unlearning dataset. By generating these fictitious biographies, we ensure that $\boldsymbol{\theta}^{\text{pre}}$ has no prior exposure to this content before undergoing LLM Surgery. To assess the effectiveness of the LLM Surgery framework in incorporating this new information, we generate 2,400 MCQA examples based on the update set and used them for evaluating $\boldsymbol{\theta}^{\text{surgery}}$'s performance. 

\paragraph{Retain Dataset ($\mathcal{D}^{\text{rtn}}$):} To ensure that the modifications to the model do not degrade the performance on unrelated tasks, we create a retain dataset comprising 2 million tokens from the OpenWebText dataset~\cite{openwebtext} and 1 billion tokens from the RedPajama-v1 dataset~\cite{rpv1}. While both $\boldsymbol{\theta}^{\text{pre}}$ and $\boldsymbol{\theta}^{\text{baseline}}$ utilize the \textit{entire} retain dataset, $\boldsymbol{\theta}^{\text{surgery}}$ only uses a \textit{small} subset of $\mathcal{D}^{\text{rtn}}$ to preserve the model's original performance on unrelated tasks. To assess the effectiveness of the LLM Surgery framework in retaining performance, we developed an evaluation benchmark consisting of 20,000 MCQA examples from the 2 million tokens in the OpenWebText dataset to ensure that $\boldsymbol{\theta}^{\text{surgery}}$’s performance remains consistent with $\boldsymbol{\theta}^{\text{pre}}$. Additionally, we evaluate $\boldsymbol{\theta}^{\text{surgery}}$ on the ARC-Easy~\cite{arc_easy} benchmark to assess its ability to retain performance comparable to \textit{off-the-shelf} Llama2-7B model.

\section{Results}
\label{sec: results}

\begin{table}[htbp]
\centering
\caption{Comparison of accuracy between $\boldsymbol{\theta}^{\text{pre}}$ and $\boldsymbol{\theta}^{\text{surgery}}$ across the unlearn, update, retain datasets, and the Arc Easy benchmark. The table also includes results from ablation studies: (1) $\boldsymbol{\theta}^{\text{gd}}$ denotes LLM Surgery with only gradient descent (gd), and (2) $\boldsymbol{\theta}^{\text{gd + kl}}$ denotes LLM Surgery with both gradient descent and KL-divergence (kl).}
\label{tab:results}
\resizebox{0.5\textwidth}{!}
{
\begin{tabular}{ccccc}
\toprule
Method & Unlearn & Update & Retain & Arc Easy \\
\cmidrule(lr){1-1} \cmidrule(lr){2-2}  \cmidrule(lr){3-3} \cmidrule(lr){4-4} \cmidrule(lr){5-5}
Llama2-7B            & 45.81   & 46.55  & 46.78  & 60.81    \\
$\boldsymbol{\theta}^{\text{pre}}$                 & 60.86   & 44.57  & 63.02  & 56.93    \\
$\boldsymbol{\theta}^{\text{surgery}}$          & 41.86   & 63.32  & 60.4   & 51.94    \\
\midrule
\multicolumn{5}{c}{Ablation Studies} \\
\midrule
$\boldsymbol{\theta}^{\text{gd}}$                   & 51.23   & 66.78  & 55.31  & 54.65    \\
$\boldsymbol{\theta}^{\text{gd + kl}}$              & 62.26   & 61.35  & 63.15  & 56.33    \\
\bottomrule
\end{tabular}
}
\end{table}

\paragraph{Effectiveness of LLM Surgery} In~\autoref{tab:results}, we empirically demonstrate the effectiveness of the LLM Surgery framework by analyzing the accuracy of the continually-pretrained Llama2-7B model across the unlearn, update, and retain evaluation benchmarks, both before ($\boldsymbol{\theta}^{\text{pre}}$) and after ($\boldsymbol{\theta}^{\text{surgery}}$) applying LLM Surgery. The results indicate that $\boldsymbol{\theta}^{\text{surgery}}$ shows a 19\% reduction in accuracy compared to $\boldsymbol{\theta}^{\text{pre}}$ on the unlearn benchmark. Additionally, on the unlearn benchmark, $\boldsymbol{\theta}^{\text{surgery}}$ exhibits 4\% lower accuracy compared to the \textit{off-the-shelf} Llama2-7B model, further confirming that $\boldsymbol{\theta}^{\text{surgery}}$ has successfully unlearned the targeted information. Simultaneously, we observe that $\boldsymbol{\theta}^{\text{surgery}}$ shows a 20\% improvement in accuracy on the update benchmark compared to $\boldsymbol{\theta}^{\text{pre}}$, demonstrating LLM Surgery's ability to incorporate new knowledge. Crucially, $\boldsymbol{\theta}^{\text{surgery}}$ maintains comparable performance on the retain and ARC-Easy benchmarks relative to $\boldsymbol{\theta}^{\text{pre}}$, suggesting that the LLM Surgery does not degrade the model’s knowledge of unrelated tasks.

\paragraph{Efficiency of LLM Surgery} To assess the efficiency gains of the LLM Surgery framework, we compare its runtime to the naive approach used to obtain the baseline model, $\boldsymbol{\theta}^{\text{baseline}}$. Empirically, we observe that $\boldsymbol{\theta}^{\text{surgery}}$ achieves the same level of performance as $\boldsymbol{\theta}^{\text{baseline}}$, but with a 35x reduction in GPU hours. This significant speed-up is due to the fact that, in practice, the unlearn and update datasets are much smaller than the retain dataset. Unlike the naive approach of retraining from scratch ($\boldsymbol{\theta}^{\text{baseline}}$) using the \textit{entire} retain dataset, LLM Surgery uses only 2\% of the retain dataset, aided by the KL divergence term in the proposed objective function. These efficiency gains become even more pronounced as the size of the retain dataset increases, highlighting the scalability and effectiveness of the LLM Surgery framework for larger datasets. Next, to evaluate the contribution of each term in the LLM Surgery objective function (see~\autoref{eqn:obj}), we conducted ablation studies. 

\paragraph{Ablation Study 1: LLM Surgery with Gradient Descent Only ($\boldsymbol{\theta}^{\text{gd}}$)} We remove the gradient ascent and KL-divergence terms, applying only the gradient descent term. As expected, relying solely on gradient descent for the update dataset improved accuracy of $\boldsymbol{\theta}^{\text{gd}}$ on the update benchmark from 44.6\% to 66.8\%, compared to $\boldsymbol{\theta}^{\text{pre}}$ (shown in~\autoref{tab:results}). However, this approach resulted in $\boldsymbol{\theta}^{\text{gd}}$ having a 10\% higher accuracy on the unlearn benchmark compared to $\boldsymbol{\theta}^{\text{surgery}}$, indicating insufficient unlearning. Moreover, accuracy on the retain set dropped by 8\% relative to $\boldsymbol{\theta}^{\text{pre}}$, suggesting that using gradient descent alone can lead to undesirable forgetting of previously learned tasks.

\paragraph{Ablation Study 2: LLM Surgery with Gradient Descent and KL-Divergence ($\boldsymbol{\theta}^{\text{gd + kl}}$)} To address the forgetting issue on $\mathcal{D}^{\text{rtn}}$, in this ablation, we include both the gradient descent and KL-divergence terms, omitting the gradient ascent term. This approach preserves the performance of $\boldsymbol{\theta}^{\text{gd + kl}}$ on $\mathcal{D}^{\text{rtn}}$, compared to $\boldsymbol{\theta}^{\text{pre}}$. However, it fails to unlearn the targeted information, as shown by the lack of accuracy drop on the unlearn benchmark. The influence of the KL-divergence term likely prevents effective unlearning. This outcome is undesirable, as $\boldsymbol{\theta}^{\text{gd + kl}}$ retains the information it was meant to unlearn.
In other words, the gradient ascent term is essential for effective unlearning, as it enables $\boldsymbol{\theta}^{\text{surgery}}$ to achieve a 19\% reduction in accuracy on the unlearn benchmark, compared to $\boldsymbol{\theta}^{\text{pre}}$, proving that all three terms are necessary for efficiently and effectively modifying LLM behavior.

\begin{figure}[htbp]
    \centering
    \includegraphics[width=\textwidth]{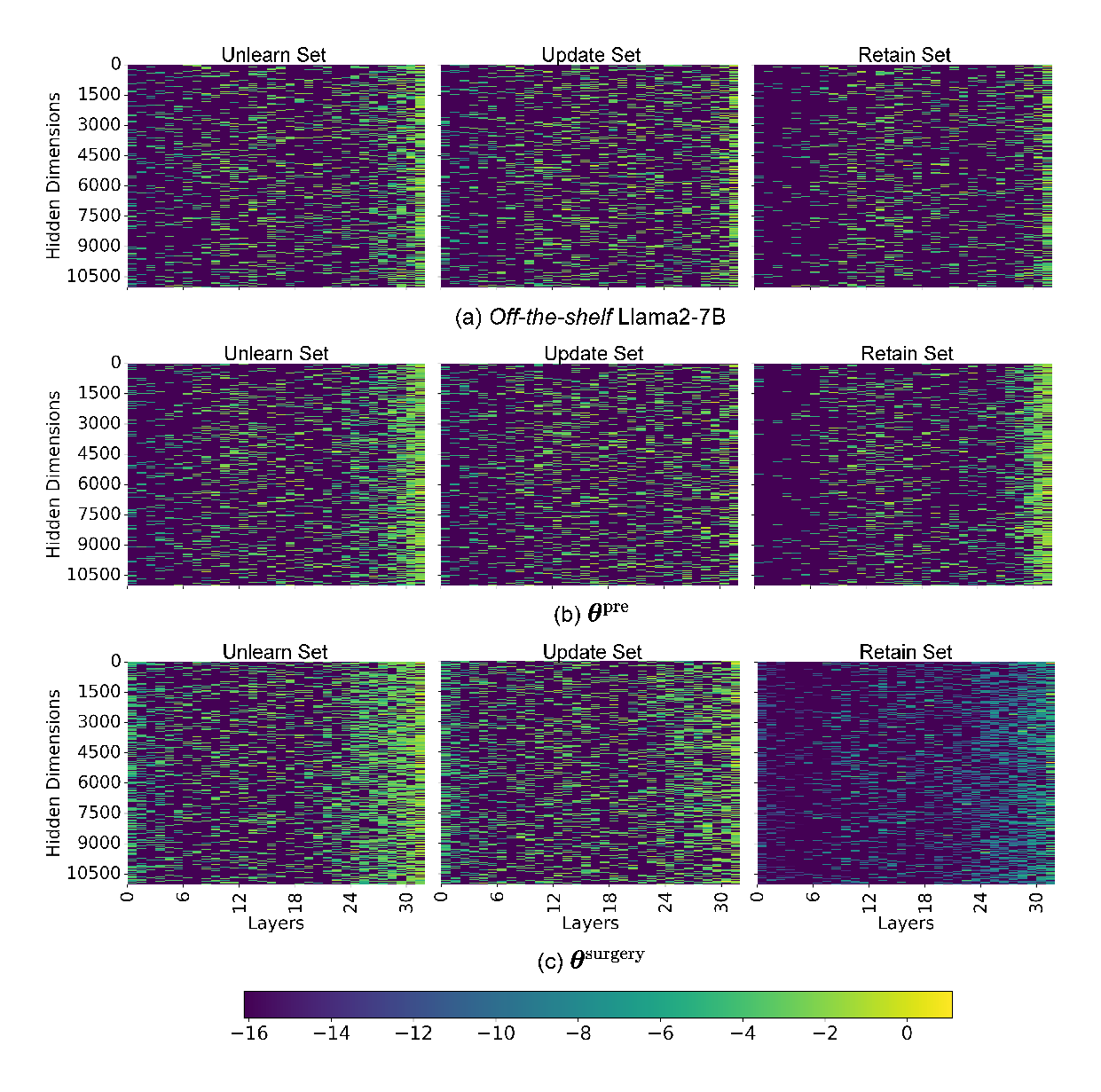}
    \caption{Visualization of activation patterns across three datasets—$\mathcal{D}^{\text{unl}}$, $\mathcal{D}^{\text{upd}}$, and $\mathcal{D}^{\text{rtn}}$—for three different models. The top row (a) shows the activation patterns for \textit{off-the-shelf} Llama2-7B, the middle row (b) represents $\boldsymbol{\theta}^{\text{pre}}$, and the bottom row (c) represents $\boldsymbol{\theta}^{\text{surgery}}$.}
  \label{fig:activations}
\end{figure}

\paragraph{Activation Pattern Analysis:}As shown in~\autoref{fig:activations}, the activation patterns remain consistent across the three datasets—$\mathcal{D}^{\text{unl}}$, $\mathcal{D}^{\text{upd}}$, and $\mathcal{D}^{\text{rtn}}$—for both the \textit{off-the-shelf} Llama2-7B and $\boldsymbol{\theta}^{\text{pre}}$. This is likely because Llama2-7B was pretrained on original Wikipedia biographies ($\mathcal{D}^{\text{unl}}$), as well as OpenWebText and RedPajama-v1 ($\mathcal{D}^{\text{rtn}}$). The same knowledge is reinforced in $\boldsymbol{\theta}^{\text{pre}}$ prior to LLM Surgery, resulting in highly similar activation patterns for \textit{off-the-shelf} Llama2-7B and $\boldsymbol{\theta}^{\text{pre}}$. After LLM Surgery, however, $\boldsymbol{\theta}^{\text{surgery}}$ shows distinct activation patterns across $\mathcal{D}^{\text{unl}}$ and $\mathcal{D}^{\text{upd}}$, as LLM Surgery modifies $\boldsymbol{\theta}^{\text{pre}}$ to unlearn $\mathcal{D}^{\text{unl}}$ and incorporate $\mathcal{D}^{\text{upd}}$, while preserving its performance on $\mathcal{D}^{\text{rtn}}$. As a result, the activation patterns on $\mathcal{D}^{\text{rtn}}$ remain similar between $\boldsymbol{\theta}^{\text{pre}}$ and $\boldsymbol{\theta}^{\text{surgery}}$.

\section{Related Works}
Several methods have been proposed for unlearning in LLMs~\cite{unlearning_survery1, unlearning_survery2, harrypotter, efficient_unlearning, tofu_benchmark, guardrail}. One notable approach~\cite{unlearning_ga_gd_kl} shares our goal of removing harmful or hallucinated content but relies on costly fine-tuning with extensive input-output annotations and requires an additional annotated dataset to ensure coherence post-unlearning, limiting scalability. Another research direction focuses on editing structured knowledge by modifying model weights~\cite{stable_editing, mend, dpo_editing}, adjusting architectures~\cite{aging_with_grace}, or localizing knowledge~\cite{kn, rome, memit, unstructured_editing}. However, these approaches are typically effective for only a small number of concepts~\cite{editing_pitfalls1, editing_pitfalls2} and fail to address unlearning of outdated information, leaving models vulnerable to information extraction attacks~\cite{attack1}.

In contrast, our method unifies unlearning and editing, enabling the removal of problematic data while updating the model. We bypass costly annotations with a continual-pretraining objective, making our approach scalable to unstructured data and large token sets. Our work also addresses regulatory requirements like the Right to be Forgotten Act~\cite{rtbf}, which mandates direct modifications to LLM weights, a limitation of in-context learning and fine-tuning alternatives.

Existing datasets fail to fully assess unlearning methods. For example,~\cite{aging_with_grace} aims to mitigate hallucinations by editing Wikipedia biographies, but pretrained knowledge likely skews the results. To address this, we introduce a new dataset, where update data is unseen by the model before editing, enabling more accurate evaluations and supporting unified approaches to unlearning and model updates.

\section{Conclusion}
We propose LLM Surgery, an efficient framework for unstructured knowledge unlearning and editing in LLMs without retraining from scratch. The proposed LLM Surgery optimization function: \textbf{(1)} Performs reverse gradient on the unlearning dataset (problematic and outdated information), \textbf{(2)} Performs gradient descent on the update dataset (new and updated information), and \textbf{(3)} Minimizes the KL divergence on the retain dataset (small subset of unchanged text), ensuring alignment between the pretrained and modified model outputs. Due to the lack of publicly available datasets specifically tailored for our novel task, we compiled a new dataset and an evaluation benchmark. Using Llama2-7B, we demonstrate that LLM Surgery can achieve significant forgetting on the unlearn set, a 20\% increase in accuracy on the update set, and maintain performance on the retain set.

\newpage

\bibliographystyle{plain}
\bibliography{main}


\end{document}